# Exploiting Transliterated Words for Finding Similarity in Inter-Language News Articles using Machine Learning

Sameea Naeem
MS Data Science
Bahria University Islamabad

Dr. Arif ur Rahman
Associate Prof / HOD CS Department
Bahria University Islamabad

Syed Mujtaba Haider
MS Data Science
Bahria University Islamabad

Abdul Basit Mughal
MS Data Science
Bahria University Islamabad

**Abstract**:
Finding similarities between two inter-language news articles is a challenging problem of Natural Language Processing (NLP). It is difficult to find similar news articles in a different language other than the native language of user, there is a need for a Machine Learning based automatic system to find the similarity between two inter-language news articles. In this article, we propose a Machine Learning model with the combination of English Urdu word transliteration which will show whether the English news article is similar to the Urdu news article or not. The existing approaches to find similarities has a major drawback when the archives contain articles of low-resourced languages like Urdu along with English news article. The existing approaches to find similarities has drawback when the archives contain low-resourced languages like Urdu along with English news articles. We used lexicon to link Urdu and English news articles. As Urdu language processing applications like machine translation, text to speech, etc are unable to handle English text at the same time so this research proposed technique to find similarities in English and Urdu news articles based on transliteration.

## 1 Introduction:

The world is now evolving around technology every researcher is trying to simplify the interaction between humans and computers. Natural language processing systems are the best to solve problems related to this field. Language is the only tool for any human being to convey their message to others and the most powerful tool of communication, but it will only be possible if the language of both the person is the same [1]. Language can be a spoken sound and words. In this world, every species like animals, humans, etc. have their language. Similarly, if we categorize language in human beings there are billions of language categories. Every human has their native language. But the problem is people belong to different cultures, countries, etc. so they might have different languages. If they want to communicate, they should have the same language, or they must understand other languages as well. As in this era, the world is evolving around technology so many problems like this are now solved by some technology. The present era is of advanced technologies with billions of machine learning solutions. For machine learning algorithms, it is not difficult to learn billions of languages. Urdu language developed with the influence of Persian, Turkish, and Hindi languages. According to google Urdu characters have some shapes like the first letter of the word, isolated, middle, and last letter of the word it is difficult to match those characters with characters of any other language. As information becomes available online and rapid increase in demographics of the internet population and abundance of Multi-language content on the web is increased, we have seen more and more published articles on every news (local news or international news) in different languages across the internet or a particular news casting platforms like Geo, BBC, and CNN, etc. It is difficult for a person to find a similar news article published in a different language on different web pages. To solve this issue this research proposes a technique to find similarities between two inter-language (Urdu and English) news articles by exploiting transliterated words and with the help of deep learning and advanced Information retrieval approaches. Finding similarities between two texts is a difficult task transliteration made it easier, it is the process of converting text from one language to another language without changing its actual meaning. For example, "lift" is a very frequent word used in the English language and in Urdu it is written as لفٹ. The objective of this article is to find similarities between two inter-language (English and Urdu) news articles using transliterated words and natural language processing (NLP) techniques.

## 2 Literature Review:

Methods like word embedding, LSTM structure, word similarities using transliteration, Term Frequency-Inverse document frequency (TF-IDF), SVM and K-means based models, and neural network-based models are discussed in some of the previous research but unfortunately, those are for other than Urdu language news articles. Other than the length of the text methods many other methods have recently been implemented to find similarities between the two texts. Therefore, a deep literature review is conducted to find English-Urdu article similarity based on transliteration and machine learning algorithms.

### 2.1 Transliteration:

It is a difficult task to write and read words that are from other languages. Transliteration is the process of converting text/words from one language to another language without changing their literal meaning. Some of the English-Urdu transliterated words are shown in below table.

| English | Urdu |
|---|---|
| break | بریک |
| train | ٹرین |
| watching | واچنگ |
| applications | ایپلیکیشنز |
| status | سٹیٹس |
| site | سائٹ |
| profile | پروفائل |
| text | ٹیکسٹ |
| boy | بوائے |
| girl | گرل |

*Figure 1: English-Urdu Transliterated words*

One of the major challenges in transliteration systems is handling missing diacritics marks or problems like space insertion between two Urdu characters. Lehal et al used machine learning techniques to made multiple language model generator to transliterate Hindi-Urdu text. The model was based on three stages [3]. Uni gram, word root founding, inspection the presence of merged words and transliterate it. The authors demonstrated research on quantifying the use of English words in Urdu news stories by exploiting transliterated words in which news stories were taken from well-known sources then feature extraction and tokenization task performed on news stories and then it is added to metadata to build archive [4]. A study proposed a technique of automatic back transliteration with probabilistic methods [2]. There are other methods of performing transliteration tasks that are demonstrated in which is consist of character alignment and decision tree learning for English-Korean, English-Japanese, English-Chinese languages. In English to Korean transliteration, most of the researchers used loan words which are used to match Korean words to English words. Transliteration rules were introduced for each English alphabet and back transliteration rule for each Korean alphabet [5]. Transliteration is known for the method of describing the word from one script to another script such as Urdu to English. Mostly rule-based approaches are used to map words of one language to another language. Research uses rule-based 20 words of dictionary which are commonly used in daily life conversations [6]. Another rule-based technique Kurdish text transliterated system used character mapping method [7]. Ahmadi et al research first converted source text to Unicode, then detection of double usage characters, then normalize the text and then characters mapping [8]. Attention based encoder-decoder model has been used in Arabic transliteration by building a transliteration corpus with parallel named entity collection, candidate selection and scoring of Arabic-English words [9]. Figure 2 shows some of the words in English-Arabic transliteration.

| Entity class | English | Arabic |
|---|---|---|
| PERSON | Villalon | فيالون |
| LOCATION | Nampa | نامبا |
| ORGANIZATION | Soogrim | سووجريم |

**Figure 2: English-Arabic Transliteration**

There are three kinds of transliteration, direct transliteration, phoneme-based transliteration, and dictionary lockup-based transliteration. Le et al proposed a technique to find transliterated words by using dictionary-based approach for English -Kashmiri language transliteration [15].

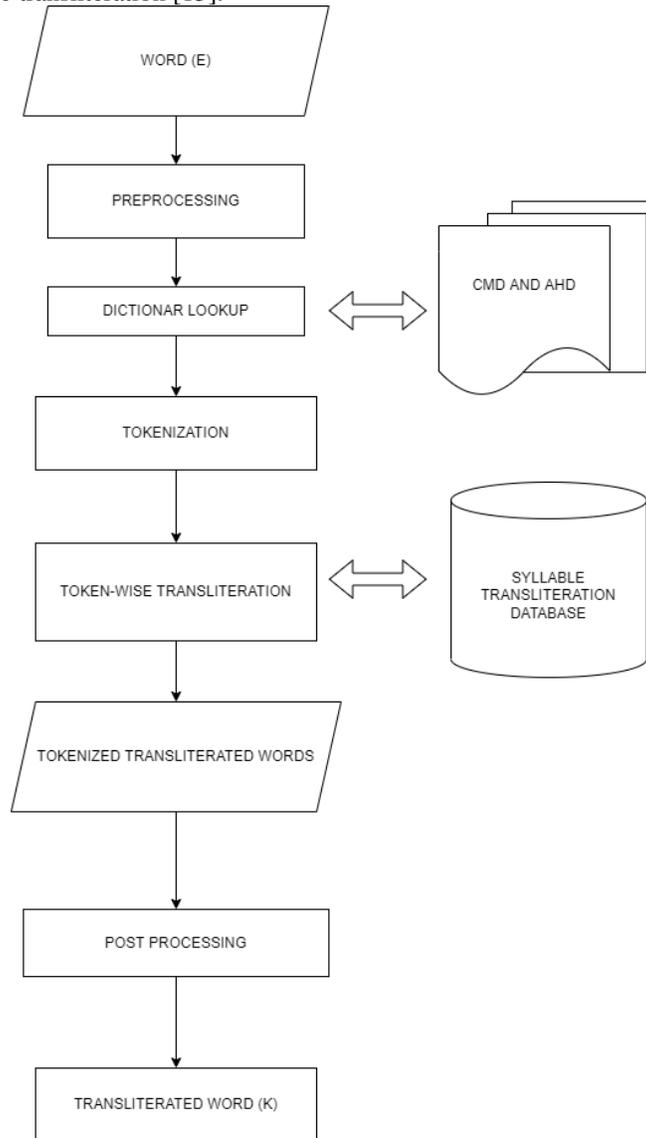

*Figure 3: Dictionary based approach*

Transliteration can be considered as a sub task of machine translation two main type of transliterations are forward transliteration and backward transliteration [10]. If we talk about rich resource language corpus building approaches are good enough but they are failed to work on low resourced languages. This kind of techniques mostly used for

English to Hindi, Korean, Chinese, Arabic transliteration. For low resourced languages a model proposed which consists of steps like preprocessed text in normalized form, removing cases, extracting alignment output from language dictionary, extracting input sequence based on output alignment, and then applied Recurrent neural networks. Transliteration can also be used in detecting news [11]. Research is consisting of some language pairs and each language pair represents a language. Shared task approach used for transliteration each pair of shared tasks consist of source or we can say a target language with specifying the transliteration direction, backward or forward. Four things were tested here word accuracy, fuzziness, mean reciprocal rank and mean average precision to find transliteration. Another research proposed technique of using recurrent neural network (RNN) along with LSTM for transliteration system of English Chinese language [12]. Many of the researchers have done transliteration on English Chinese language. One of the research projects is on the transliteration of proper nouns [13]. Mainly this research made up few steps like phonemic representation of English names for festival speech synthetic system. English phoneme sequence is then translated into generalized sequence of GIFs. GIFs represent commonly used syllabic Chinese pronunciations. Next step is to convert those GIFs into sequence of pin-yin symbols then pin-yin sequence translated into sequence of characters.

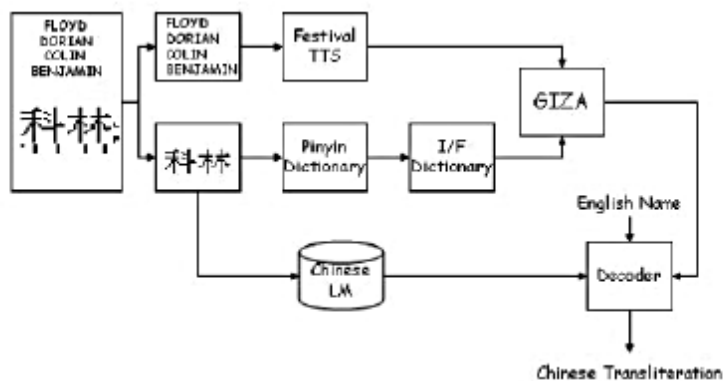

*Figure 4: English Chinese transliteration*

### 2.2 Tokenization of Urdu Text:

In every machine learning model or natural language processing model tokenization of text is the basic pre-processing step. This task is challenging in language like Urdu, Arabic, and Persian etc. It is easy to tokenize the sentence by just splitting up with spaces but it's difficult where space between words is not present. For those language models should have set boundaries where space should be inserted. When we write Urdu text by hand, we can easily put spaces where we must as we know the sequence of Urdu characters, but when Urdu text is computerized space is use occasionally according to the behavior and nature of Urdu language characters. There are two types of problems in Urdu text tokenization, one is space inclusion issues and another one is space exclusion issues. Rehman et al proposed a model for solving these issues. They use ruled-based methods by initializing joiners and non-joiner Urdu characters [14]. There are many tokenization techniques such as rule-based techniques, statistical techniques, fuzzy techniques, lexical techniques, and feature-based techniques. For avoiding space insertion issues in tokenization Zia et al present a model in which categories were defined to detect the word or whether the model should insert space or not like Affixation which will keep separate affixes from stem [15].

### 2.3 Urdu Text Boundaries:

Plenty of work is done previously in natural language processing (NLP) for western languages if we talk about Asian languages and eastern languages fewer work is done. Urdu language is widely spoken language in south Asia [16]. Due to availability of less resources Urdu is still neglected when it comes to Word2vec, tokenization, and dictionaries etc. one of the types of research worked on Urdu language processing, this research compare Urdu and Hindi language in means of tokenization, sentence boundary detection, and stop words removal etc. Stemming is the process of finding the root node of the word for example in English Girls is a word and girl is the root similarly books is the word and book is its root word. This research uses rule-based stemming for root node detection of Urdu words as shown figure 5. Stemming helps in finding nature of the word. It is also very useful for defining text boundaries of Urdu language.

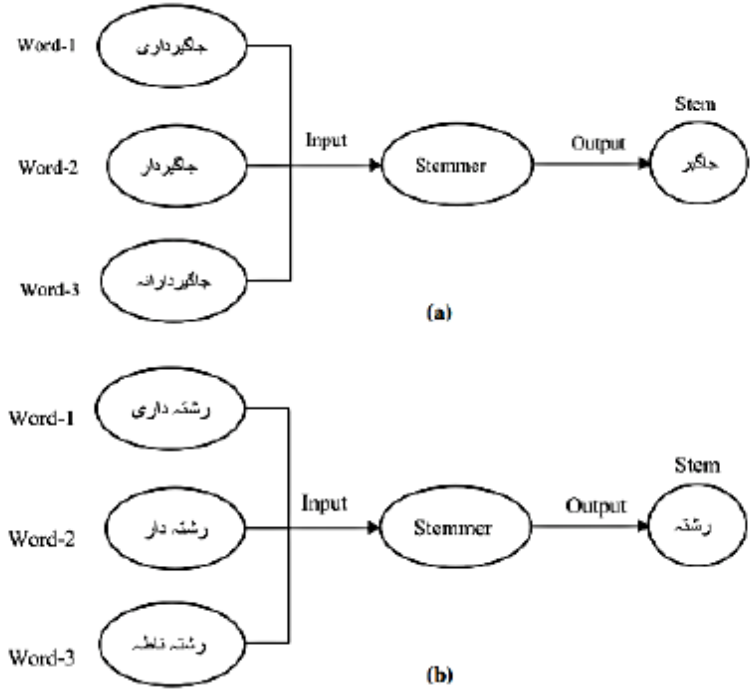

*Figure 5: Stemming for finding nature of the Urdu word.*

Word segmentation of Urdu language text is a big challenge while doing this research. For Urdu text boundaries space insertion and word segmentation is an area of constant focus. Rabiya et al presents research on Urdu word segmentation [17]. For finding text boundaries research used rule-based technique. Set of joiners and non-joiner Urdu characters were initialized. The technique used in the research includes following steps, at first step tokenization of clustered words have been done then it goes to valid word checker which will evaluate whether the tokenized word is an actual word or not, then it goes to space omission module which is responsible of applying morphology rules then the last step is space insertion between two words.

### 3 Methodology:

There are many techniques used to find similarities in inter-language news articles, many studies demonstrate types of text similarity approaches string-based similarity, corpus-based similarity, knowledge-based similarity, and hybrid similarity measures, but unfortunately very few of them are implemented yet and discussed in previous research for different languages other than English-Urdu. Finding similarity between English and Urdu news article is a difficult task but taking advantage of our daily life Urdu language which has the mixed words of English language, transliteration task is effective here. By deep literature survey on this topic this research proposed a technique which is consist of rule-based model with data normalization, word tokenization, word count scoring and Cosine similarity for English-Urdu transliteration to find similarity between two news articles. Transliteration is the process of converting a word from one language to another language without changing its literal meaning. There are so many English words which we use in our daily life Urdu language those words are known as English transliterated words.

First step is to collect inter-language (English - Urdu) documents from popular news web portals. Convert them into normalized form for example if they are not available in .txt format then first convert both to .txt normalized form. Then on second step the collected data converted into data set all the labelled entities. English transliterated word detection process will perform on Urdu news document. For transliteration process rule-based dictionary is created to match English-Urdu words. An archive data set is also used in transliteration process which is demonstrated by [4]. Data corpus built to generate two lexicons Urdu and English. Urdu lexicons will contain all Urdu words in corpus, second one will contain transliterated English words written in Urdu scripts. These lexicons will then compare with each word of Urdu document and detect transliterated words. Detection of transliterated words basically depends on Natural language processing (NLP) techniques like tokenization, POS tagging word embedding techniques along with Cosine similarity or continuous bag of words models will be performed on Urdu text news document. Tokenization is performed for English article and Urdu articles both. Part of speech (POS) tagging is the process of marking those words in a text which corresponds to some part of speech. It is the very common method of text mining. Word embedding is the learned representation of text where words that have the same meaning have a similar representation. Techniques like Cosine similarity is also commonly used in similarity findings between two texts. It is the technique which is used to find most related words for a particular word in a document. For this process word count score will be calculated for each feature extraction technique done on Urdu news article and after that weighted sum of the whole document will be calculated. Cosine similarity is a statistical approach which evaluates how relevant a word is to a document in collection documents. Proposed approach mainly uses ranking based evaluation. As the aim of this project is to find out inter-language similar news articles and suggest them to users, when the user provides Urdu news article as input, this research will compare it with English news article and make the system pick up top 10 most similar Urdu alternatives during evaluation process.

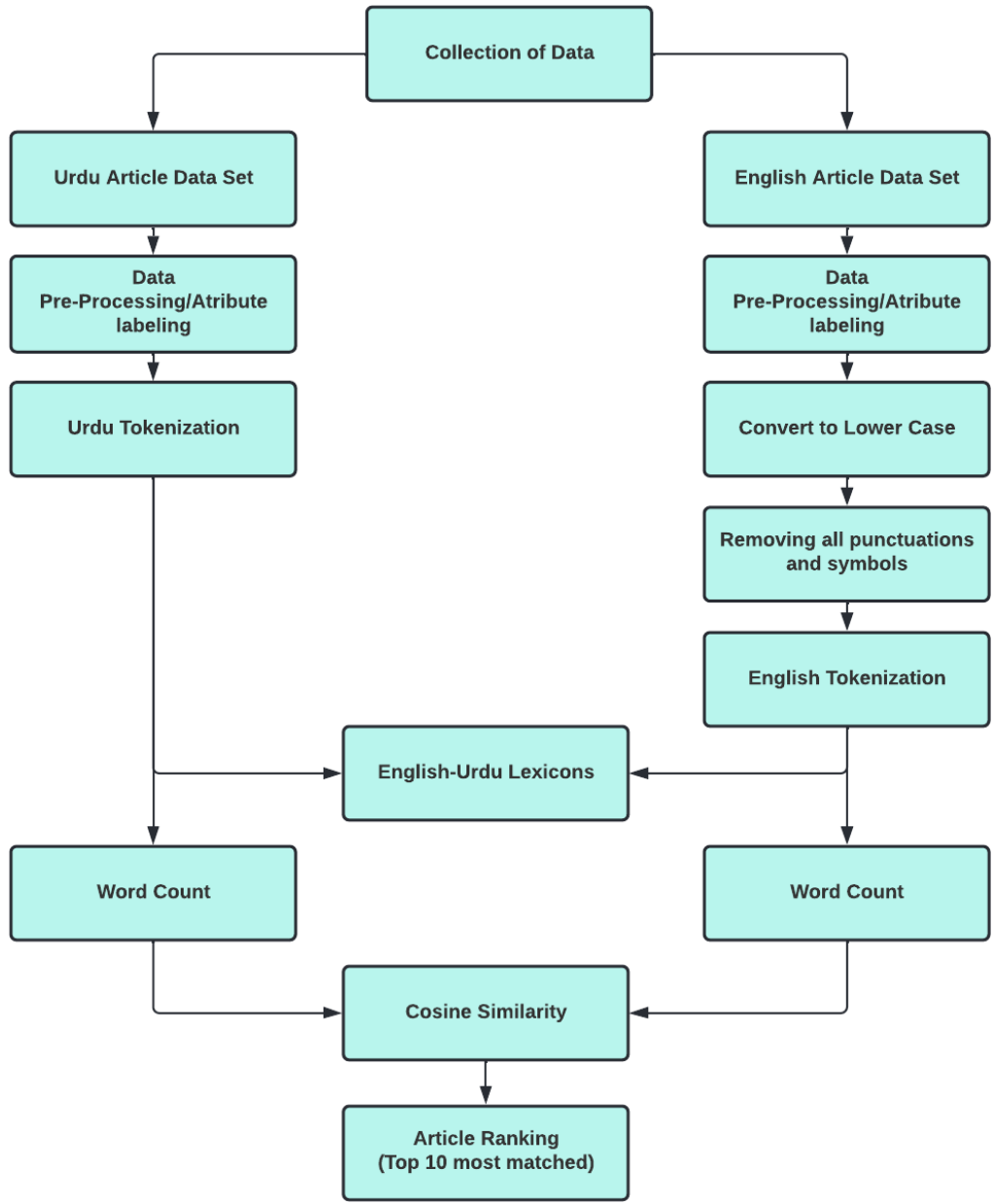

*Figure 6: Methodology*

### 3.1 Dataset Collection:

Data collection process is the process of gathering all the necessary data for model preparation or measuring the information for entities of the interest which are required for developing a model. Usually, the results and accuracy of the model depends on the attributes of collected data set. As Urdu is counted as low resourced language, very few research worked on Urdu data. No such data set available on internet, so the data collected from different news platforms like DAWN news through web crawler. Web crawler or scraper use different bots to extract data from different websites. It downloaded all the Urdu news data of different dates into a file which then converted to csv format. Data set formed was of 6 categories education, business, technology, politics, entertainment, and sports. Formed data set include 1 million Urdu news documents of DAWN news Urdu, later for quick testing, model is tested on chunk of 3000 Urdu news articles. Initially English data set contain one article downloaded from DAWN news English. This article will then compare to all the Urdu news articles present in Urdu data set for similarity matching.

### 3.2 Data Pre-processing:

Data Pre-processing is the process of manipulating and dropping of the unnecessary data for the particular model. Data pre-processing step is responsible of ensuring enhancement in processing of the model. In exploiting transliterated data to find similarities between English-Urdu news articles data pre-processing is done to make sure all the selected attributes will enhance the model results. As the data set collected was in raw form, in pre-processing selected attributes of Urdu language news articles are Date published of news article, headline, body text, source of the news article, categories of news articles, URL and news length. Similarly same attribute selected is performed on English news article. One column added to both the data sets which is index. Each paper in Urdu language articles data set and English language data set is representing an index number. In Urdu data set there are 1 million Urdu articles each article presents an index number. Initially researched was tested on a chunk of 3000 articles which means from index 0 to 2999. Similarly in English article data set one English news article is pre-processed as currently this research is based on the similarity of two interlanguage news articles.

### 3.3 Text Normalization:

Text normalization is the major step in natural language processing tasks. It is the process of normalize text on the pre-defined standards. Text normalization is responsible for input should be consistent before algorithm of the model perform operations on it. In this technique two types of text normalization are performed one is lower case conversion and removing all punctuation marks and symbols. These two are major tasks of natural language processing (NLP) in text normalization. As Urdu articles does not need to be normalized in context to these techniques, only word tokenization is enough for them therefore both are applied to English language news text. Mainly text normalization is done the news text or body text of the document.

### 3.4 Tokenization:

After data normalization text tokenization is the challenging step for Urdu Language articles. This task is easier by using NLTK libraries. As in English news articles it is easy to tokenize the text because all the characters are aligned, and no space insertion problem can occur. But for Urdu language articles space insertion is the biggest issue. As Urdu language has therefore, many different styles of characters so it is difficult to apply simple tokenization on it. For these kinds of languages different tokenization techniques are used. In this research rule-based technique is used. For space insertion issue, Urdu characters are categories into to categories first one is joiners (characters which can join) and non-joiners.

| Joiners | ب پ ت ٹ ث ج چ ح خ س ش ص ض ط ظ ع غ ف ق ک گ ل م ن ہ ی |
|---|---|
| Non-Joiners | ا د ڈ ذ ر ڑ ز ژ و ء ے |

*Figure 7: English Language News Article*

Model is able to detect the joiners which can join the next character, and non-joiners are those which cannot join, and space should be inserted between them. For avoiding these ambiguities another technique is implemented in addition with these which is well knows in previous research as well. ZWNJ is added between two non-joiner characters.

### 3.5 Word Count:

Word count also knows as word frequency, it is the process of identifying that how many times a word occurs in a document or text. In this research word count approach consist of three parts. Lexicons dictionary, English article and Urdu articles. Each word of Urdu lexicon is matched and count in English article similarly each word of English lexicon is matched and counted in Urdu articles and a list is generated. After the word count, array contains all the word count or frequency scores.

| English Lexicons | Urdu Lexicons |
|---|---|
| athlete | ایتھلیٹ |
| news | نیوز |
| videos | ویڈیوز |
| hello | ہیلو |
| fame | فیم |
| tape | ٹیپ |
| special | سپیشل |
| apartment | اپارٹمنٹ |
| role | رول |
| tweet | ٹوئٹ |
| portal | پورٹل |

*Figure 8: English-Urdu Lexicons*

### 3.6 Lexicon Building:

After deep literature review, we can say very few works is done on Urdu-English transliteration. This research follows rule-based lexicons or in simple words English-Urdu dictionary based transliteration is implemented. With the help of lexicons created by [4] this research uses 3125 English- Urdu lexicons. After deep analysis on our news articles 1234 lexicons were added to the lexicons generated by one of the previous research. Mainly this model is based on these lexicons and performing accurate transliteration. Some of the lexicons from English-Urdu lexicon dictionary are shown figure 8.

### 3.7 Cosine Similarities:

In finding similarities between two texts similarity measures are referred to the dimensional features of data object present in a data set. Cosine similarity is one of the frequently used techniques in finding cross-lingual similarity. In cosine similarity if distance measure us lesser then similarities are higher, they are inversely proportional to each other. When the distance is larger it means similarity is very low.

### 3.8 Different Similarity measures:

There are many similarity measures to find similarities between two cross lingual text similarities. Some of them are Euclidean Distance which is the most common use of distance measure. In many cases researchers refer to Euclidean distance. Euclidean distance is also known as simply distance. When data is dense or continuous, this is the best proximity measure. The Euclidean distance between two points is the length of the path connecting them. The Pythagorean Theorem gives this distance between two points. Another similarity measure is Manhattan distance in which the distance between two points is calculated as the sum of the absolute differences of their Cartesian coordinates. In a simple way of saying it is the total sum of the difference between the x-coordinates and y-coordinates. Minkowski distance is also used to calculate similarities. It is a generalized metric form of Euclidean distance and Manhattan distance. It displays results on coordinate system on x,y and z axis. The cosine similarity metric finds the normalized dot product of the two attributes. By determining the cosine similarity, we would effectively try to find the cosine of the angle between the two objects. The cosine of 0° is 1, and it is less than 1 for any other angle. Two vectors with the same orientation have a cosine similarity of 1, two vectors at 90° have a similarity of 0. Whereas two vectors diametrically opposed having a similarity of -1, independent of their magnitude. One of the reasons for the popularity of cosine similarity is that it is very efficient to evaluate, especially for sparse vectors.

### 3.9 Ranking:

After finding cosine similarity scores the last step is ranking of the articles which have higher similarity scores. The articles are arranged in ascending order depending on the scoring of cosine similarities against the index of the article in data set. This model will rank 10 most similar articles from Urdu news articles data set.

## 4   Results:

For Urdu language articles tokenization is performed using joiner and non-joiners along with ZWNJ space insertion technique which is accurately tokenize Urdu text. Tokenization is done on the body text of news articles. After tokenizing Urdu text model added a new column in the data set against each news article, that column consist of all the tokenized words.

*Figure 9: Tokenization of Urdu Articles*

After tokenization word count is done for calculating word frequency of lexicons in Urdu language data set as well as English language. Some of the rows and columns from Word count generated for Urdu article data set are shown in figure 10.

| English | Bank | funds | multi | Islamabad |
|---|---|---|---|---|
| 0 | 4 | 0 | 1 | 4 |
| 1 | 0 | 0 | 0 | 8 |
| 2 | 1 | 0 | 0 | 0 |
| 3 | 0 | 1 | 0 | 1 |
| 4 | 0 | 0 | 2 | 3 |

*Figure 10: Word Frequency Count*

After generating word count both the frequency scores are stored in a vector array. Cosine Similarity is calculated using the vector scores of each news article in Urdu data set and English data set. Cosine similarity scores are then generated against each index number. It is detecting those articles which have similarities based on the English-Urdu lexicons or transliterated words.

| | Index | Score |
|---|---|---|
| 0 | 281 | 0.606788 |
| 1 | 297 | 0.538863 |
| 2 | 440 | 0.513625 |
| 3 | 104 | 0.470391 |
| 4 | 173 | 0.444444 |
| 5 | 352 | 0.444444 |
| 6 | 393 | 0.439886 |
| 7 | 351 | 0.437193 |
| 8 | 306 | 0.414781 |
| 9 | 445 | 0.411476 |

*Figure 11: Cosine Similarity Scores*

After calculating cosine similarity score model picked 10 articles which are highly similar on the basis of cosine similarity scores and arranged them to ascending order and generate the percentage similarity of Urdu news articles with English news article and add the headline and URL of those top 10 detected Urdu news articles which are most similar to English news article on the basis of transliterated words.

10 articles are ranked from 0 to 9. Scores are in decimal form. To calculate their percentages, multiply each score by 100. Therefore, the first Urdu article which has initial score of 0.606788 with the index of 281 in the Urdu language data set is 60.63 percent similar to the English article. The next article that is similar to English article is at index 297 in Urdu data set with initial score of 0.538863 and similarity percentage of 53.89. Th third Urdu article that is similar to English article is at 440 index in Urdu data set with initial similarity score of 0.513625 with percentage of 51.36. The next similar article is at the index of 173 with initial similarity score of 0.44444 with the percentage of 44.44. The sixth similar article has same percentage of similarity and it is on the index of 352 in the Urdu language articles data set. The seventh similar Urdu article is at the index of 393 with the initial similarity of 0.439886 and the percentage is 43.99. Eighth similar article is at 351 index and percentage similarity is 43.72. Th ninth article is 41.48 percent similar and it is on the index of 306. The tenth article is on 445 index in Urdu articles data set with the percentage similarity of 41.15 percent. Limitation of the work is, currently it is finding similarities using only one English article and also using only one similarity measure which is cosine similarity. Urdu tokenization can also be done by machine learning instead of rule-based method.

| | Index | Score | Headline | URL |
|---|---|---|---|---|
| 0 | 281 | 60.68 % | بلڈرز کی وزیراعظم کو 31 کھرب روپے مدن کے تعمیرا۔ | https://www.dawnnews.tv/news/1138592/ |
| 1 | 297 | 53.89 % | ای سی سی نے زراعت ہاسنگ کے لیے 94 ارب روپے کی ... | https://www.dawnnews.tv/news/1137792/ |
| 2 | 440 | 51.36 % | او | https://www.dawnnews.tv/news/1133168/ |
| 3 | 104 | 47.04 % | غیرحتمی ڈٹ رپورٹس شائع کرنے پر پاور ڈویژن کا ا... | https://www.dawnnews.tv/news/1145294/ |
| 4 | 173 | 44.44 % | ..حکومت کا پٹرول کی موجودہ قیمت برقرار رکھنے کا۔ | https://www.dawnnews.tv/news/1143214/ |
| 5 | 352 | 44.44 % | .سونے کی فی تولہ قیمت ایک لاکھ ہزار 002 روپے تک۔۔ | https://www.dawnnews.tv/news/1135769/ |
| 6 | 393 | 43.99 % | xt... ائل کمپنیوں نے بیوروکریسی کو پٹرول بحران کا ذ | https://www.dawnnews.tv/news/1134403/ |
| 7 | 351 | 43.72 % | .بجلی کی طلب کو پورا کرنے کیلئے فرنس آئل کی درمد۔۔ | https://www.dawnnews.tv/news/1135819/ |
| 8 | 306 | 41.48 % | ایف بی میں اعلی سطح پر تقرر تبادلے | https://www.dawnnews.tv/news/1137288/ |
| 9 | 445 | 41.15 % | حکومت نے پٹرول کی قیمت مزید روپے کم کردی | https://www.dawnnews.tv/news/1132876/ |

*Figure 12: Ranking of 10 Most Similar Urdu Articles*

### 4.1 Objective and Subjective Analysis:

For finding similarities between two interlanguage news articles the best way of evaluation is human evaluation of the research. Therefore model is tested in a way that group of people selected their choice of English articles from different news portals. Each person selected one English article on which model is tested one by one and the particular person evaluated the results that the similarity scores are accurate or not and mark the research as Good, Bad, Normal. Total 15 evaluation were done with one English language article and 10 Urdu language articles, selected 10 Urdu language articles from the data set because it was easy for them to approximate the similarities manually. Some the evaluation Results are shown in figure.

|  | Good | Normal | Bad |
|---|---|---|---|
| Evaluation 1 | *** | | |
| Evaluation 2 | *** | | |
| Evaluation 3 | *** | | |
| Evaluation 4 | | *** | |
| Evaluation 5 | *** | | |
| Evaluation 6 | *** | | |
| Evaluation 7 | *** | | |
| Evaluation 8 | *** | | |
| Evaluation 9 | | *** | |
| Evaluation 10 | | *** | |

*Figure 13: Objective and Subjective Analysis*

## 5   Conclusion:

For automatic detection model to find similarity between English and Urdu news articles. To solve this problem this paper proposed a technique of using transliterated words along with Cosine similarity and natural language processing techniques to find similarity between two inter-language news articles. Proposed approach mainly uses ranking based evaluation. As the aim of this study is to find out inter-language similar news articles and rank 10 most similar Urdu articles in ascending order, when the user provides Urdu news article as input, this research will compare it with English news article and make the system pick up 10 most similar Urdu alternatives during evaluation process. Urdu language is still considered as low resourced language, so the literature review was much bounded. Rule-based lexicon approach is used in this dictionary of transliterated words. Space insertion issues are tackled by initializing joiners and non-joining Urdu characters, the biggest challenge is space insertion between two Urdu language words therefore ZWNJ approach is used.